\documentclass[3p,times]{elsarticle}

\usepackage{amssymb}

\usepackage[figuresright]{rotating}
\usepackage{longtable}
\usepackage{caption}
\usepackage{subcaption}
\usepackage{hyperref}

\begin{document}

\begin{frontmatter}


\title{Stop Oversampling for Class Imbalance Learning: A   Review}


\title{Stop Oversampling for Class Imbalance Learning: A Critical Review}

\author[label1]{Ahmad B. Hassanat}
\author[label2]{Ahmad S. Tarawneh}
\author[label3]{Ghada A. Altarawneh}
\author[label4]{Abdullah Almuhaimeed}

\address[label1]{Faculty of Information Technology, Mutah University, Karak, Jordan}

\address[label2]{Dept. of Algorithms and Their Applications,
E\"{o}tv\"{o}s Lor\'{a}nd University, Budapest, Hungary}

\address[label3]{Dept. Accounting, Mutah University, Karak, Jordan}

\address[label4]{The National Centre for Genomics and Bioinformatics, King Abdulaziz City for Science and Technology, Riyadh 11442, Saudi Arabia}

\begin{abstract}

For the last two decades, oversampling has been employed to overcome the challenge of learning from imbalanced datasets. Many approaches to solving this challenge have been offered in the literature. Oversampling, on the other hand, is a concern. That is, models trained on fictitious data may fail spectacularly when put to real-world problems. The fundamental difficulty with oversampling approaches is that, given a real-life population, the synthesized samples may not truly belong to the minority class. As a result, training a classifier on these samples while pretending they represent minority may result in incorrect predictions when the model is used in the real world. We analyzed a large number of oversampling methods in this paper and devised a new oversampling evaluation system based on hiding a number of majority examples and comparing them to those generated by the oversampling process. Based on our evaluation system, we  ranked all these methods based on their incorrectly generated examples for comparison. Our experiments using more than 70 oversampling methods and three imbalanced real-world datasets reveal that all oversampling methods studied generate minority samples that are most likely to be majority. Given data and methods in hand, we argue that oversampling in its current forms and methodologies is unreliable for learning from class imbalanced data and should be avoided in real-world applications.

\end{abstract}


\end{frontmatter}


\section{Introduction}
\label{intro}

When training a dataset with examples from one class greatly outnumbering those from the other, a phenomenon known as class imbalance emerges. The majority class is usually referred to as such, whereas the minority class is referred to as such. There may be more than one majority class and more than one minority class in a single dataset. The main cause of class imbalance is that classifiers trained on unequal training sets have a prediction bias, which is linked to poor performance in the minority class(es), Depending on the dataset utilized, the bias could range from a little imbalance to a severe imbalance \cite{r1,r2,r3,r4,r5}.
This problem has grown and has become a significant difficulty since the minority class is frequently of critical importance, as it represents favorable examples that are rare in nature or expensive to obtain \cite{r6}. This is true when considering contexts such as Big Data analytics \cite{r7,r8,r9,r10,r11,r12,r13}, Biometrics \cite{r14,r15,r16,hassanat2018identifying,tarawneh2018stability,al2019new1,hassanat2015new,al2019new2,hassanat2017classification}, gene profiling \cite{xu2018gated}, credit card fraud detection \cite{fiore2019using,ghatasheh2020cost}, face image retrieval \cite{fiore2019using}, content-based image retrieval \cite{tarawneh2019deep,tarawneh2020detailed}, disease detection \cite{hammad2021myocardial,fatima2017survey,alqatawneh2019statistical,aseeri2020modelling,hassanat2021simulation}, internet of things \cite{mnasri20173d,mnasri20183d,mnasri20153d,mnasri2020iot,abdallah2020genetic,abdallah2020emergent,mnasri2019new,mnasri2013multi,tlili2021multi,mnasri2017hybrid,mnasri2017comparative}, Natural Language Processing \cite{alghamdi2017experimental,hassanat2014rule}, network security \cite{hamadaqa2018highly,mulhem2018accelerometer,mars2019operator,alabadleh2018rss,aljaafreh2017fuzzy,abadleh2017step,abadleh2016construction}, image recognition \cite{hassanat2017hybridwavelet,hassanat2010color,hassanat2016color,narloch2019predicting,hassanat2018magnetic,hassanat2015colour}, Anomaly Detection \cite{al2019winning,al2018using,zuraiq2019phishing,almseidin2019phishing,abuzuraiq2020intelligent,almseidin2019fuzzy,al2020feature,almseidin2019detecting,alothman2020efficient,rawashdeh2018anomaly,alkasassbeh2018novel}, etc.
In formal terms, a supervised machine learning dataset $D$ with $n$ instances belonging to $m$ classes $C1, C2, C3,..., Cm$ is said to be a class imbalanced dataset if and only if for any $| Ci | >> | Cj |$, where $i$ and $j$ are indexes $1, 2, 3, ... ,m$.
There are several approaches to solving class imbalance problem before starting classification, such as:

\begin{itemize}
    \item More samples from the minority class(es) should be acquired from the discourse domain.
\item Changing the loss function to give the failing minority class a higher cost \cite{wang2010adjusted}. 
\item Oversampling the minority class.
\item Undersampling the majority class. 
\item Any combination of previous approaches.
\end{itemize}

Each of the aforementioned approaches has its own set of benefits and drawbacks \cite{hassanat2022rdpvr,fernandez2018learning}. Oversampling, on the other hand, is the most often used approach among them, as seen by the multitude of oversampling methods published in the last two decades. However, this does not necessarily imply that the oversampling approach is beneficial.
Oversampling approaches boost the quantity of minority-class instances by creating new ones out of thin air based only on their similarity to one or more of the minority's examples. This is troublesome since such methods may raise the likelihood of the learning process being overfitted \cite{branco2016survey,hauner2014latent,al2021new,fergus2018machine,al2021new}[71] [72] [73] [74] [75]. On paper, the overfitted synthetic datasets produce good machine learning results, however this is not always the case in practice. 
Another more critical problem of oversampling is that the fabricated examples could exist in the real world belonging to a different class, regardless of how similar it is to the minority’s examples, as we always have examples from class A that are the closest to examples from a different class B. Therefore, we argue that, even if such synthesizing generates favorable outcomes on paper, negative results can be easily obtained in practice.
The major goal of this study, in addition to reviewing a large number of oversampling methods, is to prove our counterclaim on the use of oversampling as a solution to the problem of class imbalance, which is as follows:

\textit{Oversampling in its current forms and methodologies is a misleading approach that should be avoided since it feeds the learning process with falsified instances that are pushed to be members of the minority class when they are most likely members of the majority.}

To the best of our knowledge, the only methodology for proving an oversampling method's goodness is its classification accuracy metrics after the classification of the oversampled datasets, with no tests for the validity of the synthesized instances and if they are appropriate for training a model for real-world use. Therefore, we find oversampling practitioners are pleased with their machine learning outcomes in the lab, but they should consider how much harm could be done in practice outside of the lab, particularly in medical and other vital applications. The harm is exacerbated when we realize that several of these methods have become integral parts of APIs and machine learning packages, such as Python imbalanced-learn API \cite{lemaitre2017imbalanced} and Smote-Variants API \cite{kovacs2019smote}.
We prove our counterclaim in this paper by using a number of typical oversampling methods on several benchmark datasets, concealing some of the majority examples, and then comparing the created examples to the hidden majority examples to determine if they approximately match. Finding such counter examples proves our counterclaim.

The following is the structure of this paper: The literature review of class imbalance problem is presented in the second section. The mythology of proving our counterclaim is illustrated in Section Three. And the experimental results are listed and discussed in section four.

\section{Literature review of oversampling methods}

\begin{figure}
    \centering
    \includegraphics[width=0.5\textwidth]{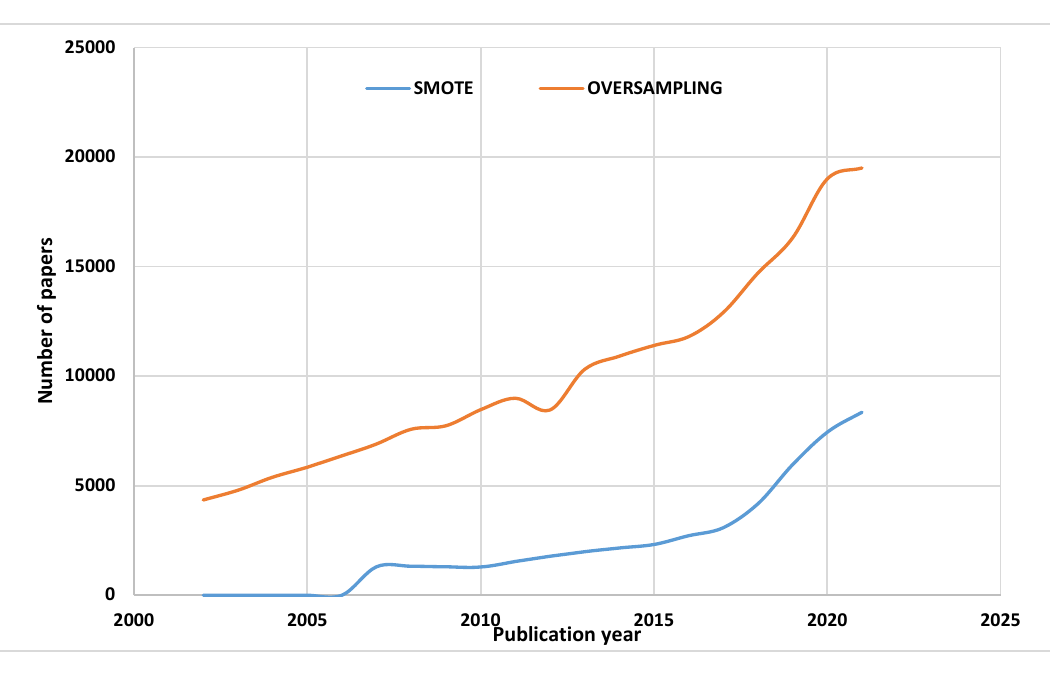}
    \caption{The number of publications that have the terms "oversampling" and/or "SMOTE"}
    \label{figpublications}
\end{figure}

In the literature, there are various ways to machine learning from class imbalance data. One of the most prevalent ways, particularly SMOTE-like approaches, is oversampling. On January 26, 2022, a Google Scholar search for the term "SMOTE" yielded 77,300 results, while a search for "oversampling" yielded 297,000 results. This is merely a foreshadowing of the developing trend of oversampling. Figure~\ref{figpublications} depicts the nearly exponential increase in the number of articles that dealt with, employed, or addressed oversampling and/or SMOTE.

The relevance of the well-defined class imbalance problem and the simplicity of oversampling solutions are the reasons for this abnormal surge in oversampling research. Anyone with a rudimentary understanding of machine learning can come up with a novel way to produce fresh similar examples given some minority examples. There could be an infinite number of such solutions.

Several studies, such as \cite{haixiang2017learning,sun2009classification,tanha2020boosting}, have reviewed various oversampling approaches; nevertheless, they are not thorough and have not paid adequate attention to validating the oversampling approach to the problem of class imbalance.

One of the earliest and most extensively utilized approaches for class imbalance is the Synthetic Minority Oversampling Technique (SMOTE) \cite{chawla2002smote}. It interpolates synthetic examples between nearest neighbors from the training set's collection of minority class cases. As a result, by merging the properties of seed instances with randomly picked k-nearest neighbors, a synthetic sample is generated. The earliest version of the SMOTE algorithm relied solely on synthetic oversampling. They also used a combination of synthetic oversampling and undersampling, which might be useful \cite{drummond2003c4}. SMOTE was tested on nine benchmark datasets and proven to improve classification performance.

SVMSMOTE \cite{tang2008svms}, which is based on SMOTE, focuses on constructing SVM modifications to successfully handle the problem of class imbalance. Oversampling, cost-sensitive learning, and undersampling are some of the heuristics used in SVM modeling. This method produced promising results when compared to other oversampling methods.

Borderline-SMOTE \cite{han2005borderline} is an SMOTE-based minority oversampling method that only oversamples the minority examples around the borderline. In comparison to SMOTE and other random oversampling methods investigated, their findings show that this solution improves classification results for the minority class.

Oversampling by a synthetic inverse minority is used in Reverse-SMOTE (R-SMOTE) \cite{das2020oversampling}, a technique based on SMOTE and the inverse near-neighbor idea. R-SMOTE beats other over-sampling methods in terms of precision, F-measurement, and accuracy, according to this study that compared traditional sampling procedures to alternative methods, including SMOTE. In the comparison, three benchmark datasets were employed.

Constrained Oversampling (CO) \cite{liu2020constrained} is a technique for reducing noise in oversampling. This method is used to extract the overlapping regions in a dataset. Ant Colony Optimization is then used to define the boundaries of minority regions. Most significantly, in order to create a balanced dataset, fresh samples are synthesized via oversampling under constraints. This method varies from others in that it includes noise-reduction constraints in the oversampling process. CO outperforms a range of oversampling benchmarks, according to their results.

In addition, the Majority Weighted Minority Oversampling Technique (MWMOTE) \cite{barua2012mwmote}  was offered as a solution to the problem of class-imbalance learning. MWMOTE finds and weights difficult-to-learn informative minority class samples based on their distance from nearby majority class samples. It then creates synthetic samples from the weighted informative minority class samples using a clustering algorithm. The primary premise of MWMOTE is that all generated samples must belong to one of the minority class clusters. In terms of numerous assessment measures, the provided results suggest that MWMOTE is superior than or similar to some other existing approaches.

Adaptive synthetic (ADASYN) \cite{he2008adasyn} was given with the goal of eliminating bias and moving the classification decision boundary in the direction of the hard examples. The primary idea behind ADASYN is to use a weighted distribution for different minority class examples based on their learning difficulty, with more synthetic data created for more difficult minority class examples than for easier minority class examples. The efficacy of this method is proved by the results of experiments conducted on a variety of datasets using five different evaluation measures.

Synthetic Minority Over-Sampling Technique Based on Furthest Neighbor Algorithm (SOMTEFUNA) \cite{r6} is another exciting and recent method for machine learning from imbalanced datasets. To produce fresh synthetic minority examples, this method employs the farthest neighbor examples. SOMTEFUNA has a number of advantages over some other approaches, one of which being the lack of tuning parameters, which makes it easier to be used in real-world scenarios. Using Naive Bayes and Support Vector Machine classifiers, the method compared the benefits of resampling to common methods such as SMOTE and ADASYN. The reported findings show that SOMTEFUNA is a viable alternative to the other oversampling methods, according to its reported results.

Sampling WIth the Majority (SWIM) \cite{bellinger2019framework} is a synthetic oversampling method that is robust in cases of significant class imbalance. SWIM's fundamental feature is that it uses the density of the well-sampled majority class to direct the creation process. SWIM's model was built using both the radial basis function and the Mahalanobis distance. SWIM was put to the test on 25 benchmark datasets, and the findings show that it beats some of the most common oversampling methods.

Other ways of oversampling include, but are not limited to, the work of \cite{tian2020new,domingos1999metacost,kurniawati2018adaptive,zhang2019wotboost,kovacs2019smote,raghuwanshi2020smote,douzas2017self,pradipta2021radius,krawczyk2019radial,hong2020oversampling,ibrahim2021odbot,wang2020multiple,bej2021loras,zhu2020improving,douzas2018improving,faris2020improving,jiang2021imbalanced,douzas2019imbalanced,zhang2018imbalanced,wang2021global,liu2018fuzzy,wu2020entropy,engelmann2020conditional,li2017boosting,wang2020improving,malhotra2019empirical,kovacs2019empirical,dhurjad2014survey,li2021novel,jiang2021new}

The validation process is what all oversampling methods have in common, which is basically the evaluation of the classifier's performance employed to classify the oversampled datasets using one or more accuracy measures such as Accuracy, Precision, Recall, F-measure, G-mean,  Specificity, Kappa, Matthews correlation coefficient (MCC), Area under the ROC Curve (AUC), True positive rate, False negative (FN), False positive (FP), True positive (TP), True negative (TN), and ROC curve. Table 1 lists 72 oversampling methods, including their known names, references, the number of datasets utilized, the number of classes in these datasets, the classifiers employed, and the performance metrics used to validate the classification results after oversampling.

\begin{center}
\begin{longtable}{lllllll}
\caption{Summary of the methods used in this study. In this table, C4.5 is Decision Tree~C4.5, LR is LR, LDA is Linear Discriminate Analysis, NB is Naive Bayes, RF is Random Forest and ANN is Artificial Neural Network.
}\\
\hline
\textbf{ID} & \textbf{Method Name} & \textbf{Reference} & \textbf{No. Datasets } & \textbf{No. Classes} & \textbf{Classifiers} & \textbf{Performance measures} \\ 
\hline
\textbf{ M1 } & SMOTE & \cite{chawla2002smote} & 6 & Binary~~ & \begin{tabular}[c]{@{}l@{}}C4.5\\Ripper\\NB\end{tabular} & \begin{tabular}[c]{@{}l@{}}ROC curve\\AUC\end{tabular} \\
\textbf{ M2 } & SMOTE TomekLinks & \cite{batista2004study} & 13 & Binary & C4.5 & AUC \\
\textbf{ M3 } & SMOTE ENN & \cite{batista2004study} & 13 & Binary & C4.5 & AUC \\
\textbf{ M4 } & Borderline SMOTE1 & \cite{han2005borderline} & 4 & Binary & C4.5 & \begin{tabular}[c]{@{}l@{}}TP rate \\F-values\end{tabular} \\
\textbf{ M5 } & Borderline SMOTE2 & \cite{han2005borderline} & 4 & Binary & C4.5 & \begin{tabular}[c]{@{}l@{}}TP rate \\F-values\end{tabular}  \\
\textbf{ M6 } & ADASYN & \cite{he2008adasyn} & 5 & Binary &  C4.5 & \begin{tabular}[c]{@{}l@{}}Accuracy, Precision\\ Recall, F-measure\\ G-mean\end{tabular} \\
\textbf{ M7 } & AHC & \cite{wang2006classification} & 1 & Binary & \begin{tabular}[c]{@{}l@{}}C4.5\\ KNN, SVM\\NB \\ AdaBoost\end{tabular} & \begin{tabular}[c]{@{}l@{}}Recall, Specificity\\Accuracy\\ ROC, G-mean \\ weighted accuracy \end{tabular} \\
\textbf{ M8 } & distance SMOTE & \cite{de2007distance} & 10 & Binary &  Linear Regression & \begin{tabular}[c]{@{}l@{}} Precision\\ Recall\\ F-measure \end{tabular} \\
\textbf{ M9 } & polynom fit SMOTE & \cite{gazzah2008new} & 1 & Binary & SVM & TP rate, TN rate \\
\textbf{ M10 } & Stefanowski & \cite{stefanowski2008selective} & 9 & Binary &\begin{tabular}[c]{@{}l@{}}   C4.5\\ MODLEM \end{tabular}  & \begin{tabular}[c]{@{}l@{}}   Accuracy\\ Specificity\\Recall \end{tabular}  \\
\textbf{ M11 } & ADOMS & \cite{tang2008generation}& 12 & Binary & ANN & G-mean \\
\textbf{ M12 } & Safe Level SMOTE & \cite{bunkhumpornpat2009safe} & 2 & Binary & NB, SVM & \begin{tabular}[c]{@{}l@{}}Precision, Recall\\F-values, AUC \end{tabular} \\
\textbf{ M13 } & MSMOTE & \cite{hu2009msmote} & 3 & Binary & \begin{tabular}[c]{@{}l@{}}C4.5\\AdaBoost\end{tabular}  & \begin{tabular}[c]{@{}l@{}}Precision, Recall\\F-values\end{tabular} \\
\textbf{ M14 } & DE oversampling & \cite{chen2010novel} & 10 & Binary & SVM & F-measure, AUC \\
\textbf{ M15 } & SMOBD & \cite{wang2012applying} & 9 & Binary & SVM & G-mean, AUC \\
\textbf{ M16 } & SUNDO & \cite{cateni2011novel} & 4 & Binary & CART, SVM & \begin{tabular}[c]{@{}l@{}}Accuracy, TP\\TN, FP, FN\end{tabular}\\
\textbf{ M17 } & MSYN & \cite{fan2011margin} & 10 & Binary & C4.5 & AUC, F-measure \\
\textbf{ M18 } & SVM balance & \cite{farquad2012preprocessing} & 1 & Binary & \begin{tabular}[c]{@{}l@{}}ANN\\ RF\\ LR \end{tabular} & \begin{tabular}[c]{@{}l@{}}Accuracy, Recall\\ Specificity, AUC \end{tabular} \\
\textbf{ M19 } & TRIM SMOTE & \cite{puntumapon2012pruning} & 11 & Binary & C4.5 & AUC, F-measure \\
\textbf{ M20 } & SMOTE RSB & \cite{ramentol2012smote} & 44 & Binary & C4.5 & AUC \\
\textbf{ M21 } & ProWSyn & \cite{barua2013prowsyn} & 10 & Binary & \begin{tabular}[c]{@{}l@{}}ANN\\C4.5 \end{tabular} & \begin{tabular}[c]{@{}l@{}}F-measure\\G-mean, AUC \end{tabular} \\
\textbf{ M22 } & SL graph SMOTE & \cite{bunkhumpornpat2013safe} & 8 & Binary & \begin{tabular}[c]{@{}l@{}}KNN, RIPPER\\C4.5\\NB \end{tabular} & F-measure, AUC \\
\textbf{ M23 } & LVQ SMOTE & \cite{nakamura2013lvq} & 8 & Binary & \begin{tabular}[c]{@{}l@{}}Logistic Tree\\ ANN\\ NB\\RF\\SVM, OLVQ3 \end{tabular} & \begin{tabular}[c]{@{}l@{}}Recall\\ Specificity\\G-mean \end{tabular} \\
\textbf{ M24 } & SOI CJ & \cite{sanchez2013synthetic} & 20 & Binary & C4.5  & \begin{tabular}[c]{@{}l@{}}Precision, Recall \\F-measure, AUC \end{tabular} \\
\textbf{ M25 } & ROSE & \cite{menardi2014training} & 20 & Binary & \begin{tabular}[c]{@{}l@{}}C4.5 \\ LR \end{tabular} & AUC \\
\textbf{ M26 } & SMOTE OUT & \cite{koto2014smote} & 18 & Binary & SVM & F-measure \\
\textbf{ M27 } & SMOTE Cosine & \cite{koto2014smote} & 18 & Binary & SVM & F-measure \\
\textbf{ M28 } & Selected SMOTE & \cite{koto2014smote} & 18 & Binary & SVM & F-measure \\
\textbf{ M29 } & LN SMOTE & \cite{maciejewski2011local} & 15 & Binary & \begin{tabular}[c]{@{}l@{}}C4.5 \\ NB \end{tabular} & \begin{tabular}[c]{@{}l@{}}F-measure \\ G-mean \\Recall \end{tabular} \\
\textbf{ M30 } & MWMOTE & \cite{barua2012mwmote} & 20 & Binary & \begin{tabular}[c]{@{}l@{}}C4.5 \\ AdaBoost, KNN \\ANN  \end{tabular} & G-mean, AUC \\
\textbf{ M31 } & PDFOS & \cite{gao2014pdfos} & 6 & Binary & RBF & \begin{tabular}[c]{@{}l@{}}F-measure \\ G-mean, AUC  \end{tabular} \\
\textbf{ M32 } & RWO sampling & \cite{zhang2014rwo} & ~ & ~ & ~ & ~ \\
\textbf{ M33 } & NEATER & \cite{almogahed2015neater} & 22 & Binary & \begin{tabular}[c]{@{}l@{}} C4.5 \\ RF, SVM\end{tabular} & G-mean, AUC \\
\textbf{ M34 } & DEAGO & \cite{bellinger2015synthetic} & 2 & Binary & ANN & AUC \\
\textbf{ M35 } & Gazzah & \cite{gazzah2015hybrid} & 2 & 100 & SVM & \begin{tabular}[c]{@{}l@{}} True Negative Rate\\ True Positive Rate \end{tabular} \\
\textbf{ M36 } & MCT & \cite{jiang2015novel} & 14 & Binary & \begin{tabular}[c]{@{}l@{}} C4.5 \\ KNN, SVM \\NB \end{tabular} & \begin{tabular}[c]{@{}l@{}} FN, FP \end{tabular} \\
\textbf{ M37 } & SMOTE IPF & \cite{saez2015smote} & 9 & Binary &   C4.5 & AUC \\
\textbf{ M38 } & KernelADASYN & \cite{tang2015kerneladasyn} & 7 & Binary & C4.5 & \begin{tabular}[c]{@{}l@{}} Precision\\ Recall\\ F-measure \\ G-Mean \end{tabular} \\
\textbf{ M39 } & MOT2LD & \cite{xie2015synthetic} & 15 & Binary & CART & \begin{tabular}[c]{@{}l@{}} G-mean \\ F-measure \end{tabular}  \\

\textbf{ M40 } & V SYNTH & \cite{young2015using} & 6 & Binary &\begin{tabular}[c]{@{}l@{}}   LR\\ LDA   \end{tabular} &  \begin{tabular}[c]{@{}l@{}}   Kappa\\ Recall \\  Specificity\\ Accuracy \\ FPR, FNR\end{tabular}\\
\textbf{ M41 } & OUPS & \cite{rivera2016priori} & 45 & Binary & \begin{tabular}[c]{@{}l@{}}   SVM\\ KNN\\ LDA\\ LR   \end{tabular}& \begin{tabular}[c]{@{}l@{}}   Recall\\ Specificity\\ G-mean  \end{tabular} \\

\textbf{ M42 } & SMOTE D & \cite{torres2016smote} & 66  & Binary & \begin{tabular}[c]{@{}l@{}}   C4.5\\ KNN, SVM   \end{tabular} & F-measure \\
\textbf{ M43 } & SMOTE PSO & \cite{cervantes2017pso} & 18 & Multi & SVM & AUC and G-mean \\
\textbf{ M44 } & CURE SMOTE & \cite{ma2017cure} & 12 & Multi & \begin{tabular}[c]{@{}l@{}}  CART\\RF \end{tabular}& \begin{tabular}[c]{@{}l@{}}  F-measure\\G-mean\\AUC \end{tabular} \\
\textbf{ M45 } & CE SMOTE & \cite{chen2010new} & 10 & Binary &  C4.5 & \begin{tabular}[c]{@{}l@{}}   F-measure \\ G-mean \end{tabular}  \\
\textbf{ M46 } & Edge Det SMOTE & \cite{kang2010weight} & 9 & Binary & SVM  & \begin{tabular}[c]{@{}l@{}}   F-measure \\ G-mean \end{tabular}\\
\textbf{ M47 } & CBSO & \cite{barua2011novel} & 8 & Binary & \begin{tabular}[c]{@{}l@{}}   ANN\\ C4.5 \end{tabular} &\begin{tabular}[c]{@{}l@{}}   accuracy\\ F-measure \\ G-mean \end{tabular} \\
\textbf{ M48 } & ASMOBD & \cite{wang2012applying} & 9 & Binary & SVM & G-mean and AUC \\
\textbf{ M49 } & Assembled SMOTE & \cite{zhou2013quasi} & 4 & Binary & SVM &\begin{tabular}[c]{@{}l@{}}   Precision\\ Recall\\ F-score\\ Accuracy \end{tabular}  \\
\textbf{ M50 } & SDSMOTE & \cite{li2014improved} & 4 & Binary & \begin{tabular}[c]{@{}l@{}}C4.5\\ AdaBoost \\ Bagging\end{tabular}& \begin{tabular}[c]{@{}l@{}}AUC \\ F-measure \end{tabular}\\
\textbf{ M51 } & DSMOTE & \cite{mahmoudi2014diversity} & 11 & Binary &  \begin{tabular}[c]{@{}l@{}}  Naïve Bayes\\ KNN, SVM \end{tabular} & \begin{tabular}[c]{@{}l@{}} Accuracy\\ Recall\\Precision\\ F-measure\\G-mean \end{tabular}\\
\textbf{ M52 } & G SMOTE & \cite{sandhan2014handling} & 4 & Binary & 1NN and SVM & AUC \\
\textbf{ M53 } & NT SMOTE & \cite{xu2014neighborhood} & 3 & Binary & SVM and C4.5 & Accuracy \\
\textbf{ M54 } & Lee & \cite{lee2015over} & 8 & Binary & SVM & \begin{tabular}[c]{@{}l@{}}Precision \\ Recall \\ G-mean \end{tabular} \\
\textbf{ M55 } & SMOTE PSOBAT & \cite{li2015optimizing} & 30 & Binary & \begin{tabular}[c]{@{}l@{}}   ANN\\ C4.5 \end{tabular} & \begin{tabular}[c]{@{}l@{}}   Kappa \\ Accuracy \end{tabular}  \\
\textbf{ M56 } & MDO & \cite{abdi2015combat1,abdi2015combat2} & 20  & Multi &

\begin{tabular}[c]{@{}l@{}}   C4.5, KNN\\ RIPPER \end{tabular} & \begin{tabular}[c]{@{}l@{}}   AUC\\G-mean\\ Precision\\ Recall\\F-measure \end{tabular} \\
\textbf{ M57 } & Random SMOTE & \cite{dong2011new} & 10 & Multi & KNN & G-mean \\
\textbf{ M58 } & VIS RST & \cite{borowska2016imbalanced} & 6 & Binary & \begin{tabular}[c]{@{}l@{}}   AdaBoost \\  C4.5 \end{tabular} & \begin{tabular}[c]{@{}l@{}}   Accuracy\\  True positive rate\\   true negatives rate\\ F-measure, and AUC \end{tabular} \\
\textbf{ M59 } & GASMOTE & \cite{jiang2016novel} & 10 & Binary & C4.5 & F-measure, G-mean \\
\textbf{ M60 } & SMOTE FRST 2T & \cite{ramentol2016fuzzy} & 1 & Binary & C4.5 & AUC, FP, FN \\
\textbf{ M61 } & AND SMOTE & \cite{yun2016automatic} & 12 & Binary & CART & AUC \\
\textbf{ M62 } & NRAS & \cite{rivera2017noise} & 41 & Binary & \begin{tabular}[c]{@{}l@{}}ANN \\ SVM, LDA \\ LR \end{tabular} & \begin{tabular}[c]{@{}l@{}}Recall \\ Specificity \\ G-mean \end{tabular} \\
\textbf{ M63 } & AMSCO & \cite{li2018adaptive} & 30 & Binary & ANN & \begin{tabular}[c]{@{}l@{}}Precision \\ Recall \\ F-measure \\ G-mean \end{tabular}\\
\textbf{ M64 } & SSO & \cite{rong2014stochastic} & 6 & Binary & ANN & \begin{tabular}[c]{@{}l@{}}  F-measure \\ G-mean \end{tabular}  \\
\textbf{ M65 } & NDO sampling & \cite{zhang2011re} & 1 & Binary & \begin{tabular}[c]{@{}l@{}}  LR \\ Decision Tree~C5 \\ ANN, SVM  \end{tabular} & F-measure, Recall \\
\textbf{ M66 } & DSRBF & \cite{fernandez2011dynamic} & 13 & Multi & ANN & \begin{tabular}[c]{@{}l@{}}   Accuracy\\ Recall  \end{tabular} \\
\textbf{ M67 } & Gaussian SMOTE & \cite{douzas2018improving} & 1 & Binary & SVM & \begin{tabular}[c]{@{}l@{}}  Accuracy\\ Recall\\Specificity\end{tabular} \\
\textbf{ M68 } & Supervised SMOTE & \cite{hu2014new} & 2 & Binary & SVM & \begin{tabular}[c]{@{}l@{}}   Accuracy\\ Recall\\Specificity\\MCC , AUC  \end{tabular} \\
\textbf{ M69 } & SN SMOTE & \cite{garcia2012surrounding} & 39 & Binary & \begin{tabular}[c]{@{}l@{}}   KNN, ANN \\ C4.5  \end{tabular} & \begin{tabular}[c]{@{}l@{}}   Accuracy\\ Recall\\Specificity, AUC  \end{tabular} \\
\textbf{ M70 } & CCR & \cite{koziarski2017ccr} & 32 & Binary & \begin{tabular}[c]{@{}l@{}}   CART, KNN\\NB, SVM   \end{tabular} & \begin{tabular}[c]{@{}l@{}}   F-measure\\ G-mean, AUC \end{tabular}  \\
\textbf{ M71 } & ANS & \cite{siriseriwan2017adaptive} & 14 & Binary & \begin{tabular}[c]{@{}l@{}}   ANN, KNN\\NB, SVM \\ C4.5    \end{tabular} & F-measure, AUC \\
\textbf{ M72 } & cluster SMOTE & \cite{cieslak2006combating} & 2 & Binary & RIPPER & AUC \\
\hline
\end{longtable}
\end{center}

As can be seen from the previous discussion and Table 1, all the aforementioned oversampling methods use the classification accuracy measures of the synthesized data to verify their goodness, assuming that the synthesised examples belong to the minority class. On paper, however, the accuracy measures appear to be good if the data is over-fitted, which is common when using Oversampling methods \cite{fernandez2018learning,branco2016survey,hauner2014latent,al2021new,fergus2018machine,hassanat2022rdpvr}.

Another critical problem with the oversampling approach is the assumption that the synthetic examples belong to the minority class; do they truly belong to the minority class?

None of the previous literature has answered this critical question. This study aims to provide a validation system for oversampling methods, in order to determine to what degree these methods synthesize unrealistic examples; assuming they are belonging to the minority when they are not.

\section{Method and Data}

The proposed validation system for oversampling methods works by hiding a subset of the majority's examples, which is referred to as the hidden subset. Although the hidden majority examples are part of the population, we excluded them from the training dataset since we assumed they were not obtained from the real-world domain of discourse. Because all oversampling approaches do not access the entire real-world population, this assumption is correct.

It is important to make sure that the class imbalance problem still exists after concealing the hidden subset.

After that, we apply the oversampling method that needs to be validated on the remaining dataset in order to generate new examples, which is referred to as the synthetic subset. The hidden subset is then returned to the training set.

The generated examples in the synthetic subset are claimed to belong to the minority class by all oversampling methods. We compare the similarity between these examples (the synthetic subset) and all examples in the original training set before oversampling to see if these synthesized examples belong to the minority or the majority.

\begin{figure}[ht!]
    \centering
    \includegraphics[width=0.8\textwidth]{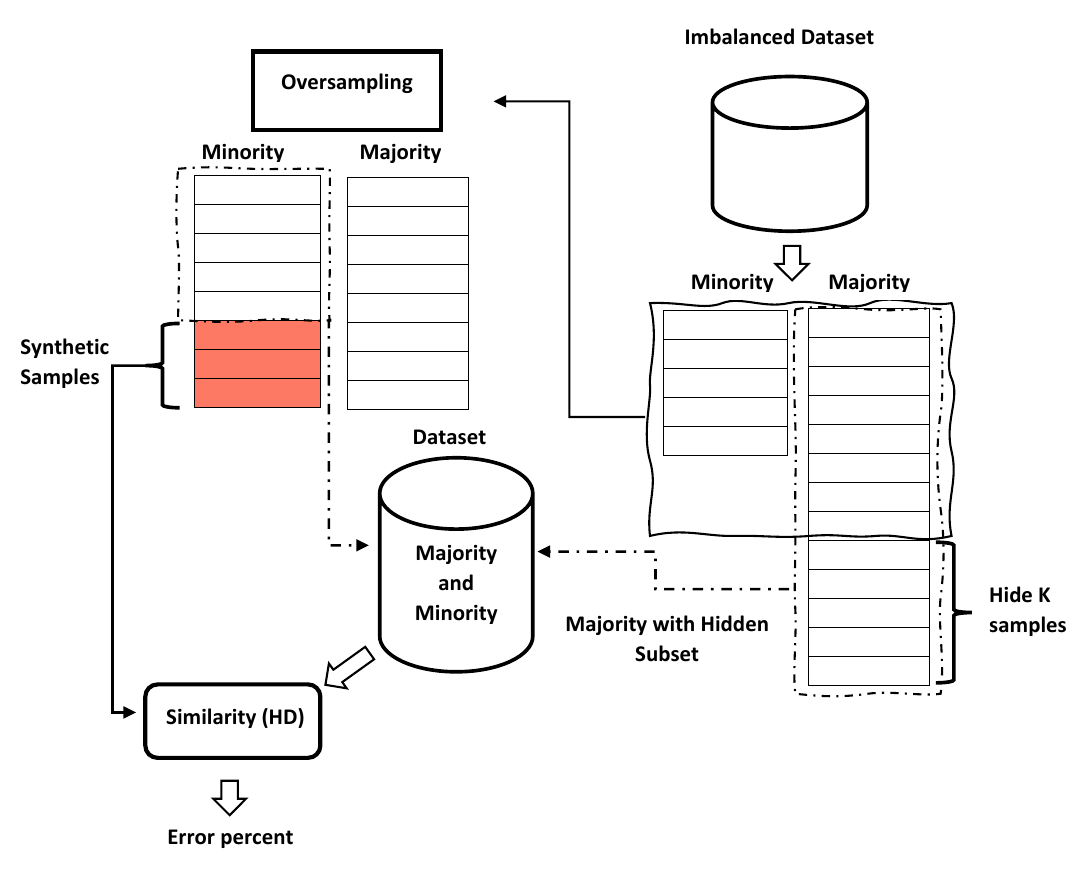}
    \caption{Flow diagram of the proposed validation system}
    \label{fig1}
\end{figure}

Figure~\ref{fig1} illustrates the proposed validation system. 

In order to determine the degree of similarity,  we need a similarity measure such as Euclidean distance (ED), Manhattan distance (MD), Hassanat distance (HD) \cite{HassantHD}, etc. In this paper, we opt for HD as being invariant to noise, outliers and data scale, since the nature of this metric prevents each feature from having a distance greater than one, regardless of the scale of the features in the targeted dataset. Furthermore, HD had been shown to outperform a wide range of machine learning similarity measures, including the most common ones like ED and MD \cite{abu2019effects, ehsani2020robust, kancharla2022latent, veerachamy2021agricultural, farooq2021computing}.




HD can be expressed mathematically as in equation\ref{eq1}.

	\begin{equation}
	D(p_i,q_i)=
	\left\{
	\begin{array}{ll}
	1-\frac{1+min(p_i,q_i)}{1+max(p_i,q_i)}  & \mbox{, } min(p_i,q_i) \geq 0 \\
	1-\frac{1+min(p_i,q_i)+|min(p_i,q_i)|}{1+max(p_i,q_i)+|min(p_i,q_i)|}  & \mbox{, } min(p_i,q_i) < 0
	\end{array}
	\right.
    \label{eq1}
	\end{equation}
\noindent	
and for the total distance between two examples is
\begin{equation}
	HD(\mathbf{p},\mathbf{q})={\sum_{i=1}^{N}D(p_i,q_i) }
    \label{eq2}
\end{equation}

\noindent Where $p$ and $q$ are feature vectors and $N$ is the number of features in each vector.

It is worth mentioning that we are proposing a validation system, not an evaluation system, the similarity measure using HD is meant to find the number of examples taken from the synthetic subset that are similar to the minority as the core of our  validation system. Those generated examples, which are more similar to the majority indicate the error of the oversampling method validated. This error is calculated according to equation~\ref{eq3}.

\begin{equation}
    Error=\frac{CM}{SS}
    \label{eq3}
\end{equation}

\noindent where $CM$ is the number of synthetic examples that are close to majority examples and $SS$ is the total number of examples in the synthetic subset.

\section{Datasets}

We employ three real-life datasets to put our validation system to the test, namely Yeast4, Yeast5, and Yeast6, which are routinely used by many oversampling methods. On \cite{Dua:2019}, all of the datasets are freely available. Table 2 contains information about these datasets.

\begin{table}[ht]
\centering
\caption{Description of the datasets used in this study}
\begin{tabular}{llllll} 
\hline
\textbf{ID} & \textbf{Name} & \textbf{No. Attributes} & \textbf{No. Classes} & \textbf{No. Minor} & \textbf{No. Major} \\ 
\hline
\textbf{1} & \textbf{Yeast4} & 10 & 2 & 51 & 1433 \\
\textbf{2} & \textbf{Yeast5} & 10 & 2 & 44 & 1440 \\
\textbf{3} & \textbf{Yeast6} & 10 & 2 & 35 & 1449 \\
\hline
\end{tabular}
\label{tbl2}
\end{table}

Table~\ref{tbl2} shows that the datasets have different minority and majority distributions, despite the fact that the number of attributes and classes are the same. It is not necessary to address the problem with multi-class datasets to prove our counter claim, as most oversampling approaches only use binary class datasets, as shown in Table 1.

\section{Experiments and Results}

In our experiments, we used all of the oversampling methods listed in Table 1 on each of the three datasets listed in Table 2, after eliminating some majority examples at random. We employed varied numbers of hidden examples, namely 10\%, 25\%, and 50\% of the majority examples of each dataset examined, to see the effect of the number of hidden examples on the validation process. 
Furthermore, each experiment is repeated five times, with the average of the results for each hidden ratio for each oversampling method on each dataset being reported. Table \ref{tbl2} shows the number of erroneous synthetic examples (NE), which are ones that are generated as minority examples but appear to be more comparable to majority examples, as the proposed validation system suggests. It also shows the number of synthetic examples (SE) generated by each oversampling method, in addition to the error rate (ER) which is calculated using Equation \ref{eq3}. All the result reported in Table \ref{tbl2} were obtained by hiding only 10\% of the majority examples.

\begin{center}
\begin{longtable}{llllllllllll}
\caption{Oversampling Methods' Validation Results on three datasets using 10\% hidden percent.}\\
\hline
                & \multicolumn{3}{c}{\textbf{Yeast4}}                                                                 & \multicolumn{3}{c}{\textbf{Yeast5}}                                                                 & \multicolumn{3}{c}{\textbf{Yeast6}}                                                                 & \multicolumn{1}{l}{}                 & \multicolumn{1}{l}{}               \\
\textbf{Method} & \multicolumn{1}{l}{\textbf{NE}} & \multicolumn{1}{l}{\textbf{SE}} & \multicolumn{1}{l}{\textbf{ER}} & \multicolumn{1}{l}{\textbf{NE}} & \multicolumn{1}{l}{\textbf{SE}} & \multicolumn{1}{l}{\textbf{ER}} & \multicolumn{1}{l}{\textbf{NE}} & \multicolumn{1}{l}{\textbf{SE}} & \multicolumn{1}{l}{\textbf{ER}} & \multicolumn{1}{l}{\textbf{Avg. ER}} & \multicolumn{1}{l}{\textbf{Avg. Rank}}  \\
\hline
\textbf{M1} & 463.2 & 1239 & 0.37 & 108 & 1252 & 0.09 & 294 & 1270 & 0.23 & 0.23 & 24 \\
\textbf{M2} & 470 & 1235 & 0.38 & 117 & 1252 & 0.09 & 279 & 1270 & 0.22 & 0.23 & 25 \\
\textbf{M3} & 414.2 & 1130.2 & 0.37 & 91 & 1222 & 0.07 & 266 & 1187 & 0.22 & 0.22 & 20 \\
\textbf{M4} & 270.6 & 1239 & 0.22 & 277 & 1252 & 0.22 & 391 & 1270 & 0.31 & 0.25 & 38 \\
\textbf{M5} & 471 & 1239 & 0.38 & 597 & 1252 & 0.48 & 757 & 1270 & 0.60 & 0.48 & 58 \\
\textbf{M6} & 568.8 & 1239 & 0.46 & 219 & 1252 & 0.17 & 534 & 1270 & 0.42 & 0.35 & 50 \\
\textbf{M7} & 50 & 50 & 1.00 & 42 & 43 & 0.98 & 34 & 34 & 1.00 & 0.99 & 71 \\
\textbf{M8} & 482.6 & 1239 & 0.39 & 145 & 1252 & 0.12 & 258 & 1270 & 0.20 & 0.24 & 28 \\
\textbf{M9} & 765 & 1224 & 0.63 & 204 & 1232 & 0.17 & 244 & 1260 & 0.19 & 0.33 & 47 \\
\textbf{M10} & 53 & 59 & 0.90 & 9 & 11 & 0.82 & 17 & 17 & 1.00 & 0.91 & 68 \\
\textbf{M11} & 523.4 & 1239 & 0.42 & 165 & 1252 & 0.13 & 372 & 1270 & 0.29 & 0.28 & 43 \\
\textbf{M12} & 109.8 & 125.2 & 0.88 & \multicolumn{1}{l}{-} & \multicolumn{1}{l}{-} & \multicolumn{1}{l}{-} & 55 & 55 & 1.00 & 0.94 & 70 \\
\textbf{M13} & 363 & 1239 & 0.29 & 261 & 1252 & 0.21 & 308 & 1270 & 0.24 & 0.25 & 37 \\
\textbf{M14} & 425.8 & 771.4 & 0.55 & \multicolumn{1}{l}{-} & \multicolumn{1}{l}{-} & \multicolumn{1}{l}{-} & 250 & 648 & 0.39 & 0.47 & 56 \\
\textbf{M15} & 479 & 1239 & 0.39 & 116 & 1252 & 0.09 & 357 & 1270 & 0.28 & 0.25 & 39 \\
\textbf{M16} & 285 & 286 & 1.00 & 289 & 292 & 0.99 & 298 & 300 & 0.99 & 0.99 & 72 \\
\textbf{M17} & 309.8 & 1239 & 0.25 & 45 & 1252 & 0.04 & 79 & 1270 & 0.06 & 0.12 & 6 \\
\textbf{M18} & 478.8 & 1239 & 0.39 & 103 & 1252 & 0.08 & 269 & 1270 & 0.21 & 0.23 & 22 \\
\textbf{M19} & 329 & 1239 & 0.27 & 26 & 1252 & 0.02 & 124 & 1270 & 0.10 & 0.13 & 7 \\
\textbf{M20} & 938.8 & 2478 & 0.38 & 0 & 43 & 0.00 & 575 & 2540 & 0.23 & 0.20 & 13 \\
\textbf{M21} & 729.4 & 1239 & 0.59 & 221 & 1252 & 0.18 & 385 & 1270 & 0.30 & 0.36 & 51 \\
\textbf{M22} & 282.6 & 1239 & 0.23 & \multicolumn{1}{l}{-} & \multicolumn{1}{l}{-} & \multicolumn{1}{l}{-} & 406 & 1270 & 0.32 & 0.27 & 42 \\
\textbf{M23} & 775.8 & 1239 & 0.63 & 330 & 1252 & 0.26 & 621 & 1270 & 0.49 & 0.46 & 54 \\
\textbf{M24} & 52 & 1239 & 0.04 & 120 & 1252 & 0.10 & 41 & 1270 & 0.03 & 0.06 & 2 \\
\textbf{M25} & 867 & 1239 & 0.70 & 456 & 1252 & 0.36 & 707 & 1270 & 0.56 & 0.54 & 61 \\
\textbf{M26} & 442 & 1239 & 0.36 & 76 & 1252 & 0.06 & 246 & 1270 & 0.19 & 0.20 & 14 \\
\textbf{M27} & 475.4 & 1239 & 0.38 & 86 & 1252 & 0.07 & 265 & 1270 & 0.21 & 0.22 & 19 \\
\textbf{M28} & 470.2 & 1239 & 0.38 & 121 & 1252 & 0.10 & 303 & 1270 & 0.24 & 0.24 & 30 \\
\textbf{M29} & 413.8 & 1239 & 0.33 & 167 & 1252 & 0.13 & 272 & 1270 & 0.21 & 0.23 & 23 \\
\textbf{M30} & 644.8 & 1239 & 0.52 & 223 & 1252 & 0.18 & 369 & 1270 & 0.29 & 0.33 & 48 \\
\textbf{M31} & 872.2 & 1239 & 0.70 & 395 & 1252 & 0.32 & 599 & 1270 & 0.47 & 0.50 & 59 \\
\textbf{M32} & 927.8 & 1239 & 0.75 & 698 & 1252 & 0.56 & 987 & 1270 & 0.78 & 0.69 & 63 \\
\textbf{M33} & 1041.6 & 2478 & 0.42 & 315 & 2504 & 0.13 & 813 & 2540 & 0.32 & 0.29 & 45 \\
\textbf{M34} & 1236 & 1239 & 1.00 & 523 & 1252 & 0.42 & 36 & 1270 & 0.03 & 0.48 & 57 \\
\textbf{M35} & 504 & 791 & 0.64 & 148 & 807 & 0.18 & 159 & 841 & 0.19 & 0.34 & 49 \\
\textbf{M36} & 602 & 1239 & 0.49 & 191 & 1252 & 0.15 & 414 & 1270 & 0.33 & 0.32 & 46 \\
\textbf{M37} & 518 & 1239 & 0.42 & 110 & 1252 & 0.09 & 292 & 1270 & 0.23 & 0.25 & 34 \\
\textbf{M38} & 1200 & 1239 & 0.97 & 855 & 1252 & 0.68 & 1264 & 1270 & 1.00 & 0.88 & 67 \\
\textbf{M39} & 307 & 1226 & 0.25 & 129 & 1248 & 0.10 & 93 & 1262 & 0.07 & 0.14 & 8 \\
\textbf{M40} & 1199 & 1239 & 0.97 & 639 & 1252 & 0.51 & 1262 & 1270 & 0.99 & 0.82 & 66 \\
\textbf{M41} & 938 & 1264 & 0.74 & 547 & 1268 & 0.43 & 711 & 1272 & 0.56 & 0.58 & 62 \\
\textbf{M42} & 1239 & 1239 & 1.00 & 1010 & 1251 & 0.81 & 1271 & 1271 & 1.00 & 0.94 & 69 \\
\textbf{M43} & 78 & 153 & 0.51 & 32 & 132 & 0.24 & 47 & 105 & 0.45 & 0.40 & 53 \\
\textbf{M44} & 288 & 1239 & 0.23 & 188 & 1252 & 0.15 & 89 & 1270 & 0.07 & 0.15 & 9 \\
\textbf{M45} & 710 & 1239 & 0.57 & 72 & 1252 & 0.06 & 139 & 1270 & 0.11 & 0.25 & 35 \\
\textbf{M46} & 440 & 1239 & 0.36 & 85 & 1252 & 0.07 & 278 & 1270 & 0.22 & 0.21 & 17 \\
\textbf{M47} & 753 & 1239 & 0.61 & 363 & 1252 & 0.29 & 633 & 1270 & 0.50 & 0.47 & 55 \\
\textbf{M48} & 302 & 1239 & 0.24 & 35 & 1252 & 0.03 & 32 & 1270 & 0.03 & 0.10 & 4 \\
\textbf{M49} & 476 & 1239 & 0.38 & 123 & 1252 & 0.10 & 308 & 1270 & 0.24 & 0.24 & 32 \\
\textbf{M50} & 451 & 1239 & 0.36 & 85 & 1252 & 0.07 & 256 & 1270 & 0.20 & 0.21 & 16 \\
\textbf{M51} & 14 & 1239 & 0.01 & 2 & 1252 & 0.00 & 0 & 1270 & 0.00 & 0.00 & 1 \\
\textbf{M52} & 534 & 1239 & 0.43 & 85 & 1252 & 0.07 & 296 & 1270 & 0.23 & 0.24 & 33 \\
\textbf{M53} & 413 & 1239 & 0.33 & 18 & 1252 & 0.01 & 285 & 1270 & 0.22 & 0.19 & 11 \\
\textbf{M54} & 452 & 1239 & 0.36 & 103 & 1252 & 0.08 & 320 & 1270 & 0.25 & 0.23 & 26 \\
\textbf{M55} & 225 & 592 & 0.38 & 76 & 869 & 0.09 & 169 & 724 & 0.23 & 0.23 & 27 \\
\textbf{M56} & 592 & 1239 & 0.48 & 453 & 1252 & 0.36 & 441 & 1270 & 0.35 & 0.40 & 52 \\
\textbf{M57} & 470 & 1239 & 0.38 & 116 & 1252 & 0.09 & 304 & 1270 & 0.24 & 0.24 & 29 \\
\textbf{M58} & 232 & 411 & 0.56 & 69 & 224 & 0.31 & 95 & 148 & 0.64 & 0.50 & 60 \\
\textbf{M59} & 64 & 165 & 0.39 & 12 & 133 & 0.09 & 18 & 116 & 0.16 & 0.21 & 15 \\
\textbf{M60} & 552 & 1482 & 0.37 & 115 & 1500 & 0.08 & 279 & 1270 & 0.22 & 0.22 & 21 \\
\textbf{M61} & 212 & 1239 & 0.17 & 73 & 1252 & 0.06 & 128 & 1270 & 0.10 & 0.11 & 5 \\
\textbf{M62} & 140 & 1239 & 0.11 & 84 & 1252 & 0.07 & 28 & 1270 & 0.02 & 0.07 & 3 \\
\textbf{M63} & 275 & 805 & 0.34 & 97 & 1101 & 0.09 & 182 & 789 & 0.23 & 0.22 & 18 \\
\textbf{M64} & 579 & 1235 & 0.47 & 236 & 1250 & 0.19 & 257 & 1270 & 0.20 & 0.29 & 44 \\
\textbf{M65} & 558 & 1239 & 0.45 & 143 & 1252 & 0.11 & 307 & 1270 & 0.24 & 0.27 & 41 \\
\textbf{M66} & 521 & 1239 & 0.42 & 101 & 1252 & 0.08 & 275 & 1270 & 0.22 & 0.24 & 31 \\
\textbf{M67} & 918 & 1239 & 0.74 & 727 & 1252 & 0.58 & 1025 & 1270 & 0.81 & 0.71 & 64 \\
\textbf{M68} & 365 & 1239 & 0.29 & 140 & 1252 & 0.11 & 100 & 1270 & 0.08 & 0.16 & 10 \\
\textbf{M69} & 496 & 1239 & 0.40 & 104 & 1252 & 0.08 & 326 & 1270 & 0.26 & 0.25 & 36 \\
\textbf{M70} & 1048 & 1240 & 0.85 & 898 & 1253 & 0.72 & 1076 & 1270 & 0.85 & 0.80 & 65 \\
\textbf{M71} & 381 & 1239 & 0.31 & \multicolumn{1}{l}{-} & \multicolumn{1}{l}{-} & \multicolumn{1}{l}{-} & 105 & 1270 & 0.08 & 0.20 & 12 \\
\textbf{M72} & 411 & 1239 & 0.33 & 46 & 1252 & 0.04 & 498 & 1270 & 0.39 & 0.25 & 40 \\
\hline
\end{longtable}
\end{center}

The averages of five trials on each dataset for each approach are provided in Table~\ref{tbl2}. In addition, for each approach, the average error is calculated for the error rates on all three datasets. The last column in the table shows the average rank of each method based on the three datasets; the lower the rank, the better the oversampling performance; for example, rank 1 shall be awarded to the method with the smallest error.

\begin{figure}[ht!]
     \centering
     \begin{subfigure}[b]{0.3\textwidth}
         \centering
         \includegraphics[width=\textwidth]{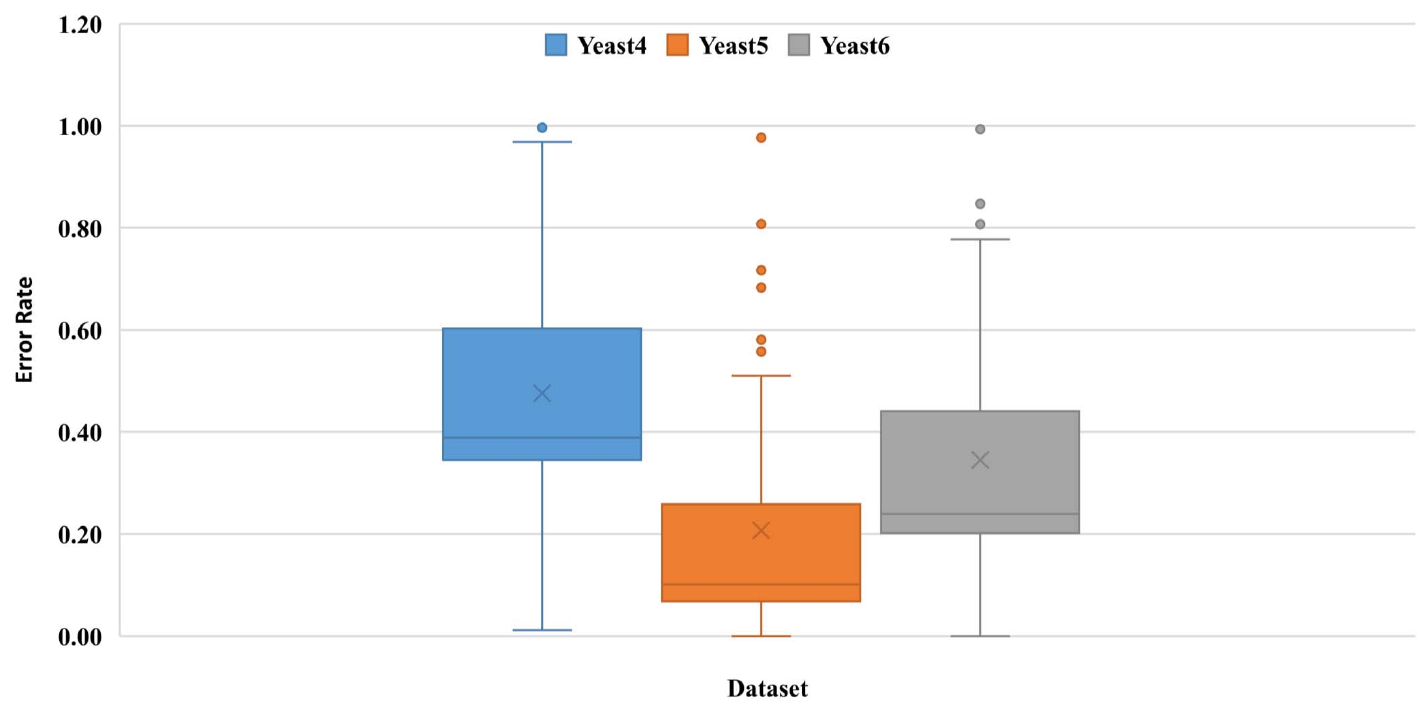}
         \caption{10\% }

     \end{subfigure}
     \hfill
     \begin{subfigure}[b]{0.3\textwidth}
         \centering
         \includegraphics[width=\textwidth]{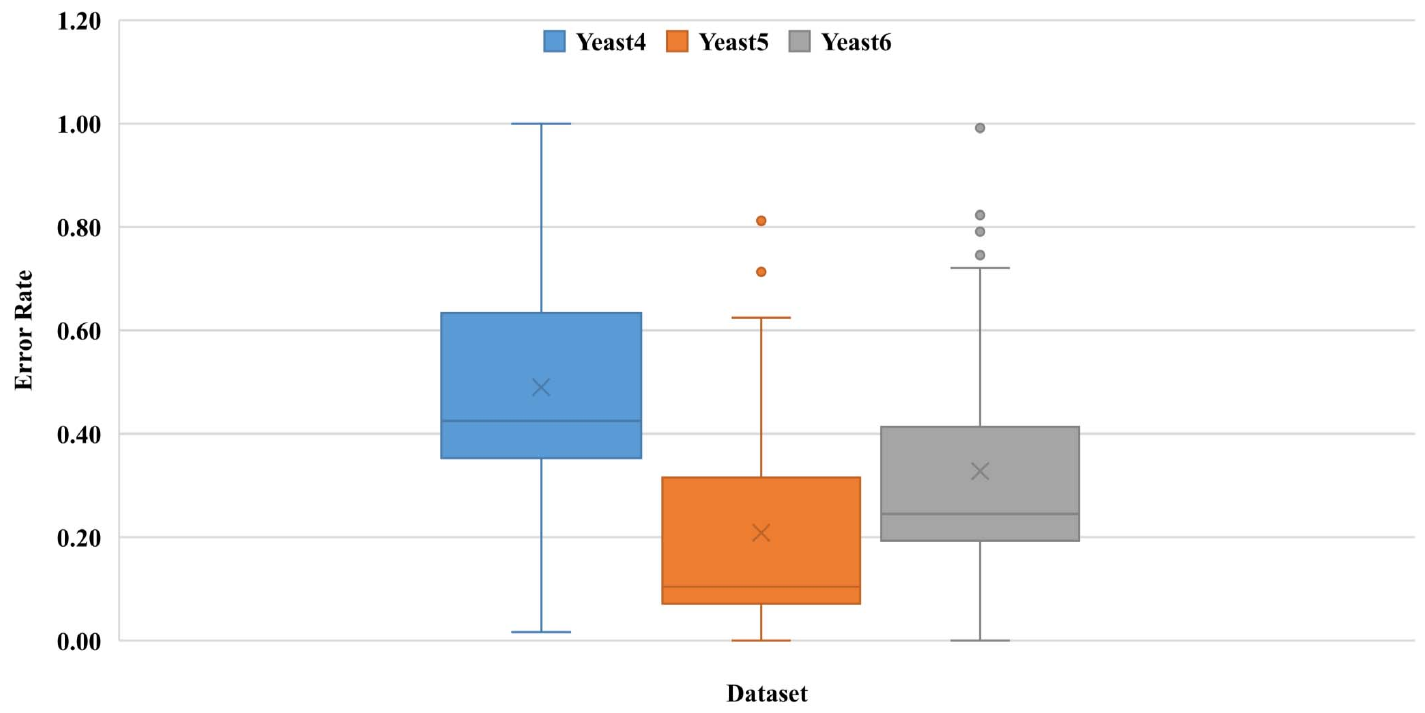}
         \caption{25\% }

     \end{subfigure}
     \hfill
     \begin{subfigure}[b]{0.3\textwidth}
         \centering
         \includegraphics[width=\textwidth]{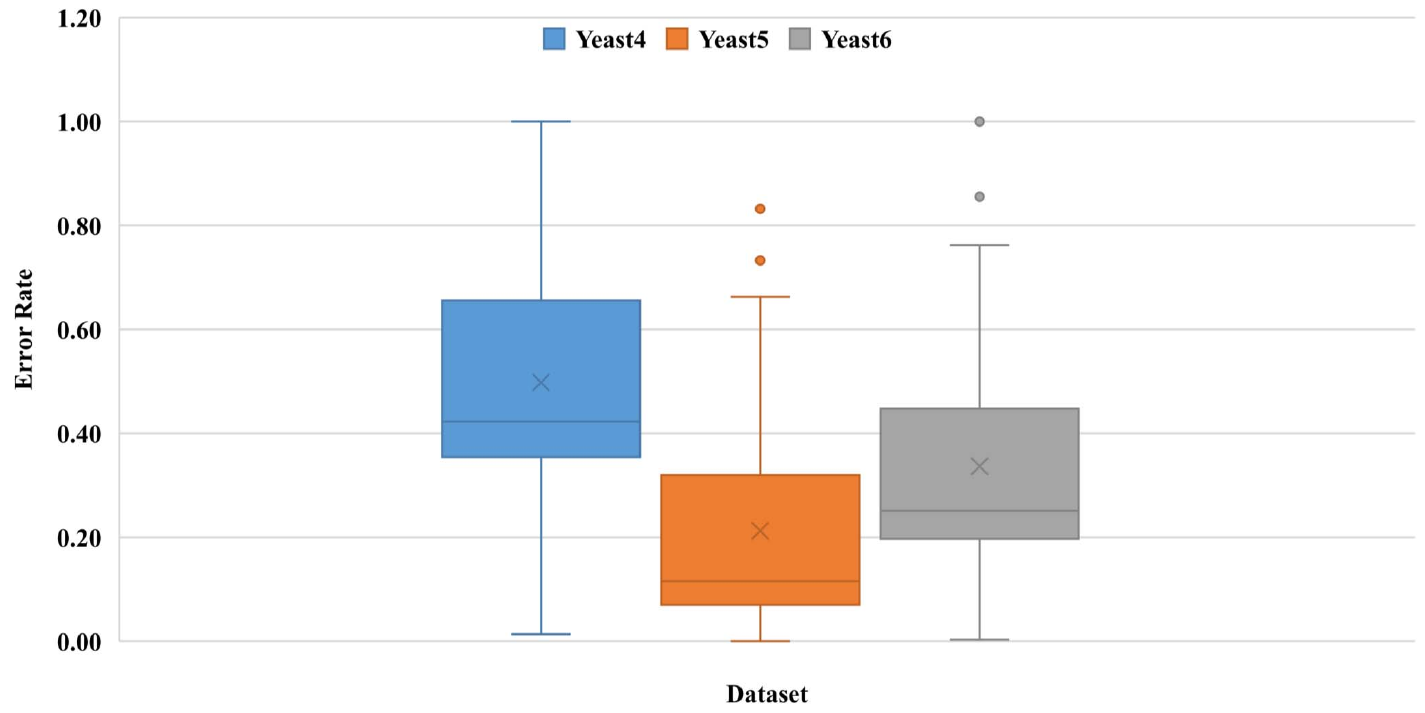}
         \caption{50\%}

     \end{subfigure}
        \caption{Box plot of the average error rates of all oversampling methods on three datasets with varied hidden percentages.}
        \label{fig3}
\end{figure}

A thorough examination of Table~\ref{tbl2} demonstrates that all oversampling methods result in errors in the synthesized examples. That is, they generate examples that are meant to be minority, yet are similar to the majority or fall within the majority class's decision boundary. Despite the fact that all methods generate such examples, the quantity of fake examples generated differs from one method to another. On the Yeast4 dataset, for example, the oversampling method (M51) generates 14 incorrect examples, whereas other methods, such as M70, generate more than 1K incorrect examples. That is why M51 is ranked first, whereas M70 is ranked much higher.
Similar findings were achieved when 25\% and 50\% of the majority examples were used as hidden examples, thus there is no need to include them in tables; however, we show them in Figure \ref{fig3}. 

The average error rate of all oversampling methods increases somewhat as the hidden percent increases, as seen in Figure \ref{fig3}. This is logical since when oversampling methods synthesize their minority examples, they become unaware of some majority examples; in fact, we expected a significant error rise as the size of the hidden subset grew larger. In terms of the effect of the dataset on the average oversampling error, we can observe in the same figure that some datasets, such as (Yeast5), are easier to be oversampled than others, such as (Yeast6) and (Yeast4). However, the difference is not substantial, and more importantly, as the Box plots show, the standard deviation of the error rates produced by all oversampling methods on each dataset is extremely high. 

\begin{figure}[ht!]
    \centering
    \includegraphics[width=\textwidth]{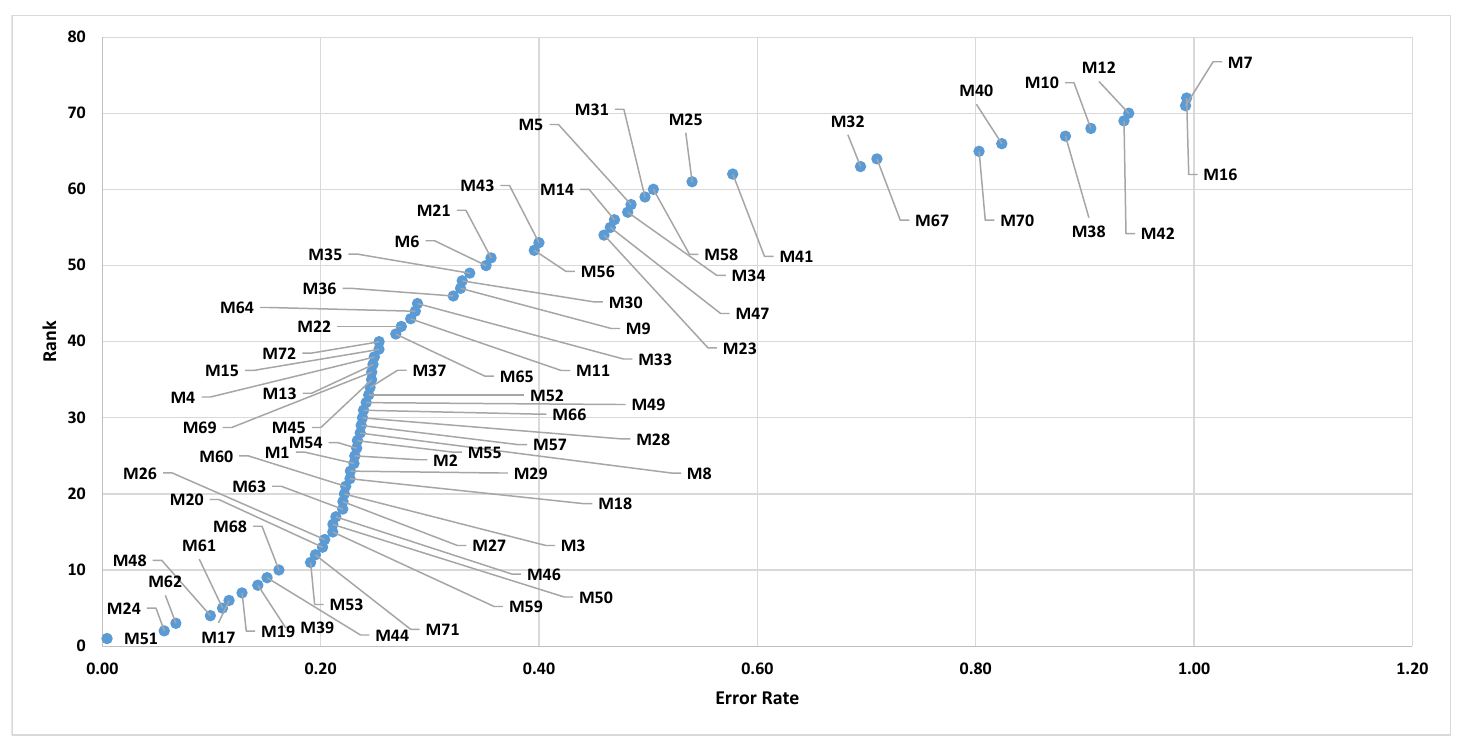}
    \caption{Methods ranking based on their average error rate on thee datasets}
    \label{fig4}
\end{figure}

In order to compare oversampling methods, we ranked them according to their average error across three datasets. The average ranks of the methods are plotted against the average errors they produced on three datasets (Yeast4, Yeast5 and Yeast6) as shown in Figure \ref{fig4}.

Despite the fact that all of the methods discussed generate false examples, Figure \ref{fig4} indicates that certain methods do better than others in avoiding the formation of false examples. As a result, if oversampling is unavoidable, the method's reliability should be verified using a validation tools such as ours. Methods like M51, for example, have error rates close to 0\%, whereas others like M7 and M16 have error rates near to 100\%.

\begin{figure}[ht!]
     \centering
     \begin{subfigure}[b]{0.3\textwidth}
         \centering
         \includegraphics[width=\textwidth]{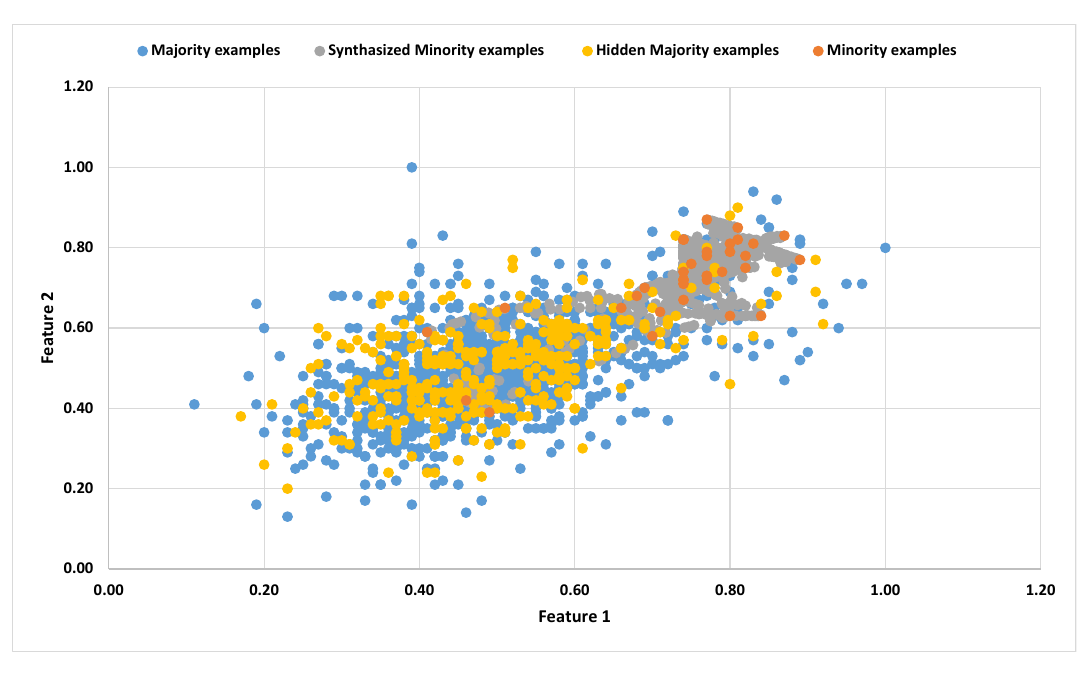}
         \caption{Yeast4 }

     \end{subfigure}
     \hfill
     \begin{subfigure}[b]{0.3\textwidth}
         \centering
         \includegraphics[width=\textwidth]{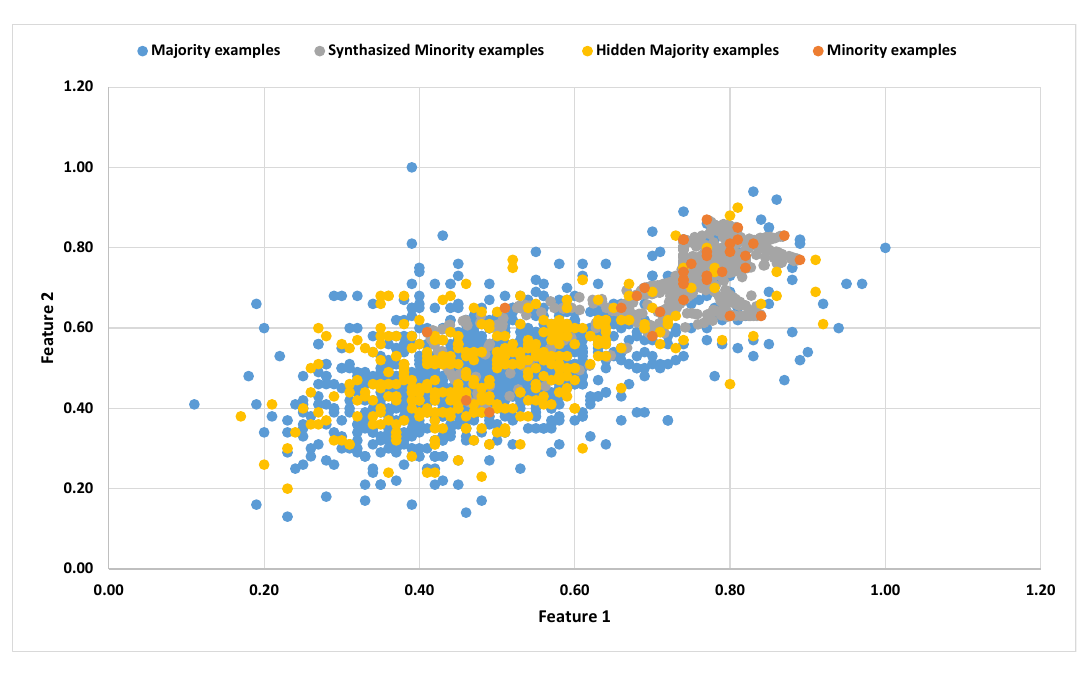}
         \caption{Yeast5 }

     \end{subfigure}
     \hfill
     \begin{subfigure}[b]{0.3\textwidth}
         \centering
         \includegraphics[width=\textwidth]{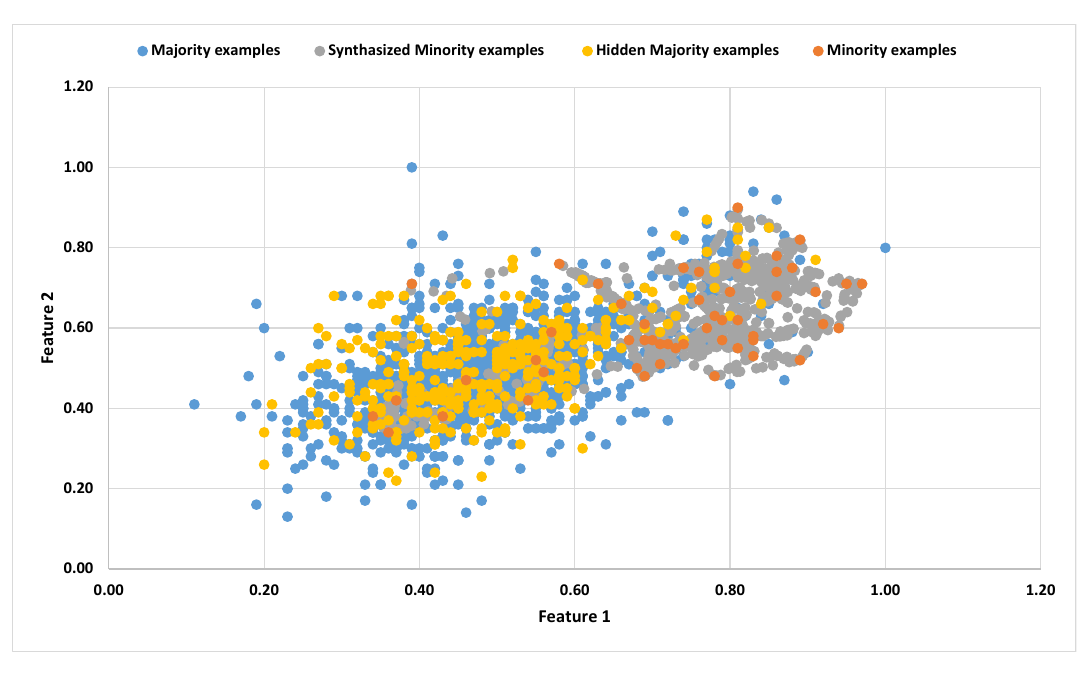}
         \caption{Yeast6}

     \end{subfigure}
        \caption{Visualization of validating the SMOTE method on three datasets. When the figure is scaled up, more information is obtained.}
        \label{fig5}
\end{figure}

As a result, all Oversampling methods validated produce misleading examples, regardless of the hidden percentage or dataset used. Figure \ref{fig5} visualizes the Oversampling results on three datasets using M1, displaying various sorts of examples, including hidden, minority, majority, and synthesized examples while hiding 10\% of the majority examples. For the sake of illustration, we limited the data to only two features.

As seen in Figure \ref{fig5}, many synthetic examples are created on the basis of or near hidden examples, producing almost identical feature values. Even in higher dimensional feature space, such a situation has the potential to occur. The common mistake that all oversampling methods make is to feed such data to a classifier, assuming that all of the examples are realistic and labeled based on reality. The classifier has no other knowledge and learns based on the false assumption, which produces excellent results in labs but unexpected behavior in real-world applications.

The results shown thus far do not necessarily imply that incorrect example synthesis occurs just when the majority examples are hidden from the oversampling method. Even though the majority examples are completely visible to the methods, some methods generate false examples. The published findings of all of the oversampling methods demonstrate this, as none of them claimed to be an accurate method with no errors.

We validated the best performers on a fourth machine learning dataset, Vehicle3, because some oversampling approaches passed our validation test by presenting a relatively small number of unrealistic examples, and to further support our counterclaim against the validity of the oversampling approach in general. The validation results of the best performers are shown in Table \ref{tblBestMethods}.

\begin{table}[ht!]
\centering
\caption{ER on Vehicle3 dataset using 25\% hidden data and some of the methods with the least ER}
\begin{tabular}{llll} 
\hline
\textbf{Method} & \textbf{NE} & \textbf{SE} & \textbf{ER} \\ 
\hline
\textbf{M51} & 168 & 264 & 0.64 \\
\textbf{M24} & 88 & 264 & 0.33 \\
\textbf{M62} & 122 & 264 & 0.46 \\
\textbf{M61} & 61 & 264 & 0.23 \\
\textbf{M19} & 91 & 264 & 0.34 \\
\textbf{M39} & 102 & 236 & 0.43 \\
\textbf{M44} & 194 & 264 & 0.73 \\
\textbf{M68} & 149 & 264 & 0.56 \\
\hline
\end{tabular}
\label{tblBestMethods}
\end{table}

As can be seen in Table \ref{tblBestMethods}, when we changed the dataset, the errors of the "best" oversampling methods increased significantly, demonstrating once again that these oversampling methods fill in the features space gap without considering whether the generated examples are truly belong to the minority, and falsely consider them as such. This makes the training of these examples deceptive, and it could lead to the classifier being overfitted on incorrect data if robust generalization techniques are not used. As a result, when applied to real-world tasks, it is possible that the entire machine learning system fails spectacularly, particularly in critical  applications such as security, autonomous driving, aviation safety and medical applications, where even one unrealistic synthesized example could do catastrophic harm.

\section{Conclusion}

Oversampling methods have been used and developed for decades to handle the problem of class imbalance learning, and there is a near exponential growing trend for such type of research. The main question of this research is oversampling approach in its current form and methods provide applicable and viable solution for learning from class imbalance data? 
We claim that the current oversampling approach is deceptive and could lead to severe failures in real-world applications. 
In order to answer the main question and to prove our counterclaim, we reviewed a large number of oversampling methods and analyzed their performance in terms of providing unrealistic examples, for this purpose we propose a new  validation system for oversampling methods, which we utilized to validate over 70 different oversampling methods. Our validation results on four real-world common datasets reveal that all of the oversampling methods investigated generate false examples, assuming that they are minorities when they are not, causing classifiers to perform well in labs but more likely fail in practice. 

The Oversampling methods investigated in this paper are ranked according to how many incorrect examples they generate. When used to solve real-life problems, the ranking shows that some methods are less harmful than others. When the datasets were changed, however, they were found to be useless. Therefore, we recommend avoiding such methods when dealing with sensitive applications such as security, autonomous driving, aviation safety, and medical applications that use machine learning from class imbalanced data. Instead, we seriously encourage using ensemble approaches to problems of class imbalance, such as Easy Ensemble. \cite{liu2008exploratory}, Random Data Partitioning \cite{hassanat2022rdpvr}, etc. Because these methods do not create data out of thin air and do not, as the Undersampling approach suggests, deny the learning process from critical data. 

More research should be done in the future to confirm the validity or invalidity of oversampling approach, investigating more methods and incorporating more data. Furthermore, we recommend that additional research be conducted on real-world applications, including measurements of incorrect predictions made with and without the use of oversampling methods, as well as comparisons with ensemble methods.

\bibliographystyle{elsarticle-num}
\bibliography{ecrc-template.bib}







\end{document}